\documentclass[10pt,twocolumn,letterpaper]{article}

\usepackage{cvpr}      

\usepackage[dvipsnames]{xcolor}

\usepackage{multirow}
\definecolor{cvprblue}{rgb}{0.21,0.49,0.74}
\usepackage[pagebackref,breaklinks,colorlinks,citecolor=cvprblue]{hyperref}

\def\paperID{25} 
\def\confName{CVPR}
\def\confYear{2024}

\title{Parameter Efficient Fine-tuning of Self-supervised ViTs \\without Catastrophic Forgetting}

\author{Reza Akbarian Bafghi$^{1}$\thanks{Joint first-authorship.}\\
\and
Nidhin Harilal$^{1}$\footnotemark[1]\\
\and
Claire Monteleoni$^{1,2}$\\
\and
Maziar Raissi$^{3}$\\
\and
$^{1}$ University of Colorado, Boulder\\
\and
$^{2}$ INRIA, Paris\\
\and
$^{3}$ University of California, Riverside\\
\and
{\tt\small
\{reza.akbarianbafghi,
nidhin.harilal,
cmontel\}@colorado.edu, maziar.raissi@ucr.edu
}
}
\begin{document}
\maketitle
\begin{abstract}
Artificial neural networks often suffer from catastrophic forgetting, where learning new concepts leads to a complete loss of previously acquired knowledge. We observe that this issue is particularly magnified in vision transformers (ViTs), where post-pre-training and fine-tuning on new tasks can significantly degrade the model's original general abilities. For instance, a DINO ViT-Base/16 pre-trained on ImageNet-1k loses over $70\%$ accuracy on ImageNet-1k after just 10 iterations of fine-tuning on CIFAR-100. Overcoming this stability-plasticity dilemma is crucial for enabling ViTs to continuously learn and adapt to new domains while preserving their initial knowledge. In this work, we study two new parameter-efficient fine-tuning strategies: (1)~Block Expansion, and (2) Low-rank adaptation (LoRA). Our experiments reveal that using either Block Expansion or LoRA on self-supervised pre-trained ViTs surpass fully fine-tuned ViTs in new domains while offering significantly greater parameter efficiency. Notably, we find that Block Expansion experiences only a minimal performance drop in the pre-training domain, thereby effectively mitigating catastrophic forgetting in pre-trained ViTs\footnote{We have made the source code available to the public at: \href{https://github.com/rezaakb/peft-vit}{\texttt{https://github.com/rezaakb/peft-vit}}.}. 
\end{abstract} 
\section{Introduction}
\label{sec:intro}
Humans excel at continual learning, gradually acquiring new information while retaining previously learned concepts, with a complete loss of prior knowledge being a rare occurrence. In contrast, artificial neural networks, including vision image transformers (ViT)~\cite{dosovitskiy2020image}, often suffer from catastrophic forgetting of old concepts as new ones are learned~\cite{goodfellow2013empirical, mccloskey1989catastrophic}. Catastrophic forgetting directly stems from the stability-plasticity dilemma, a fundamental issue in neural networks that balances the integration of new knowledge (plasticity) against the preservation of existing knowledge (stability)~\cite{mermillod2013stability}. Addressing this dilemma could enhance their applicability in dynamic, real-world scenarios. In these cases, the model must continually be fine-tuned with data from new class or domain distributions ~\cite{volpi2021continual,safaei2023open,Loaliyan_2024_CVPR,safaei2024entropic}, while preserving its prior knowledge.

We observe that fine-tuning a DINO ViT~\cite{caron2021emerging}, initially pre-trained on the ImageNet-1K~\cite{deng2009imagenet}, across various transfer datasets, results in decreased performance on the original ImageNet-1K. Notably, we find that even a short fine-tuning period of 10 iterations on CIFAR-100~\cite{Krizhevsky2009LearningML} leads to a significant drop of $ 70\%$ accuracy on the original ImageNet-1K, highlighting the pronounced effects of catastrophic forgetting in pre-trained ViTs. This underscores the need for fine-tuning approaches that preserve the model's general capabilities while integrating new, domain-specific knowledge. In response, we investigate Parameter-Efficient Fine-tuning (PEFT) strategies to focus on adjusting a minimal number of model parameters, aiming to maintain the model's foundational strengths while incorporating new insights.

PEFT strategies have been extensively explored in Natural Language Processing (NLP)~\cite{Houlsby2019Parameter,Wang2022AdaMixMF} and multimodal foundation models~\cite{Sung2021VLADAPTERPT,Zhai2023InvestigatingTC,wang2023parameter}, garnering significant attention for their ability to fine-tune models without extensively altering their structure. The exploration of these strategies has recently extended to vision models~\cite{Xin2024ParameterEfficientFF}, with research focusing on various aspects of parameter-efficient fine-tuning including efficiency~\cite{Yuan2024FullLoRAATEB,Jie2023RevisitingTP,He2022ParameterEfficientMA}, scalability~\cite{chen2022adaptformer,liang2022expediting}, transferability~\cite{pan2022st,lagunas2023transfer}, and robustness~\cite{chefer2022optimizing}. On the other hand, efforts to tackle catastrophic forgetting have been made independently~\cite{Zhang2019SideTuningAB,Lee2022SurgicalFI,Ramasesh2022EffectOS}, yet incorporating cutting-edge PEFT techniques from NLP into the vision domain to mitigate catastrophic forgetting in Vision Transformers (ViTs) remains an underexplored area. Our research distinguishes itself from these broad explorations of PEFT characteristics by specifically focusing on leveraging PEFT techniques to mitigate catastrophic forgetting during the fine-tuning of Vision Transformers (ViTs), thereby presenting a novel study in this domain.
\begin{figure*}[t]
    \centering
    \includegraphics[width=\linewidth]{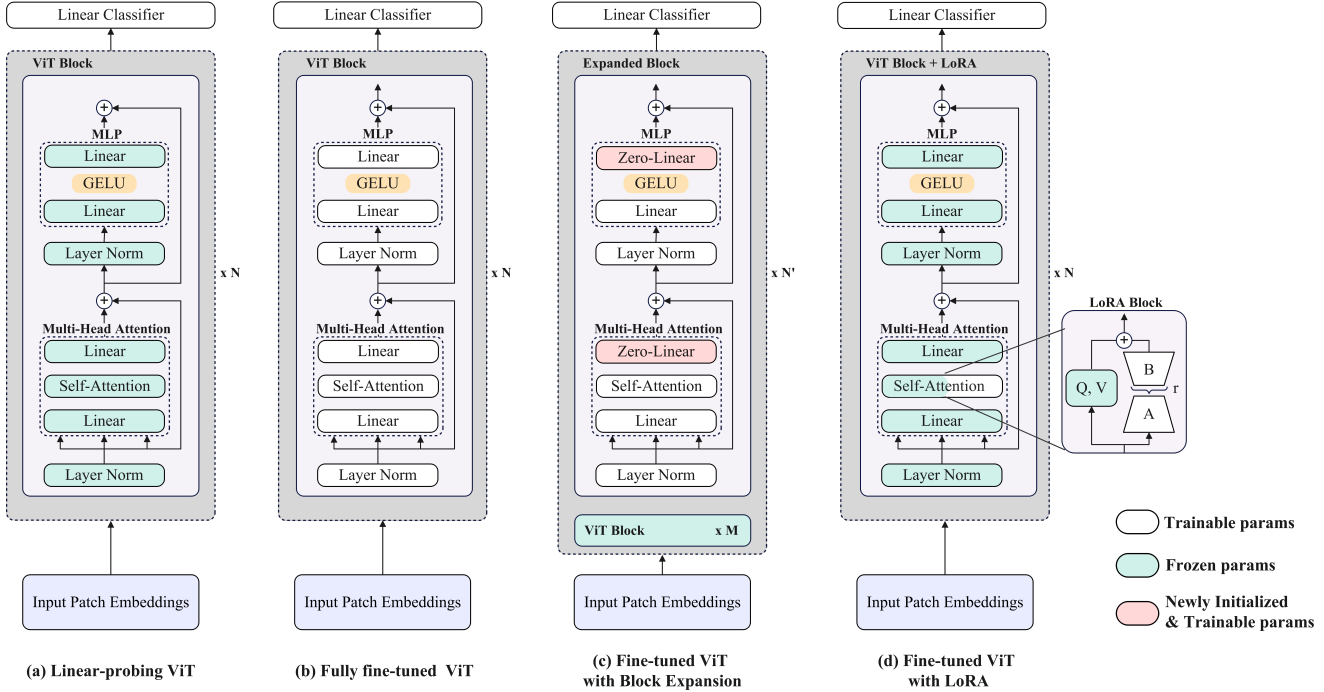}
    \caption{Comparison of different ViT fine-tuning approaches: (a) Linear-probing ViT model with all weights frozen (cyan blocks) and a trainable classifier (white block). (b) Fully fine-tuned ViT with all trainable weights (white blocks). (c) Block Expansion with additional blocks containing trainable zero-initialized linear layers (red blocks) and other trainable parameters (white blocks). (d) Low-Rank Adaptation (LoRA) weights (white blocks) added in parallel to the frozen pre-trained weights of Queries and Values (cyan blocks).}
    \label{fig:framework}
\end{figure*}
Drawing on the recent Low-Rank Adaptation (LoRA)~\cite{hu2021lora} and Block Expansion~\cite{wu2024llama} techniques from Natural Language Processing, we adapt and apply these concepts to ViTs. Our work conducts a thorough comparison of these innovative strategies against traditional fine-tuning methods, focusing particularly on their ability to retain performance on original tasks while excelling in new domains. 
Our findings reveal that both Block Expansion and LoRA exhibit a remarkable ability to adapt to diverse transfer domains. However, LoRA's performance may deteriorate in scenarios involving simpler datasets like CIFAR-10. In contrast, Block Expansion demonstrates a robust ability to safeguard the model's performance on the pre-training dataset.


\section{Method}
This section provides an overview of conventional fine-tuning approaches and delves into the application of LoRA and Block Expansion techniques to ViTs.
\label{sec:method}
\vspace{-10pt}
\paragraph{Standard Fine-tuning.}
Standard fine-tuning is a set of commonly adopted techniques to adapt pre-trained models for new tasks or domains. It starts with a pre-trained model, whose weights and biases are initialized from its original training phase. For full fine-tuning, every parameter of the model, encompassing weights and biases across all layers, is updated via gradient changes based on the new task or dataset as shown in Figure~\ref{fig:framework}. The model is considered a differentiable function, allowing gradients derived from the new task's loss to adjust all the trainable parameters. Other variations in standard fine-tuning involve: Top-k (Fine-tuning only top-k layers) and linear probing (Fine-tuning only the linear classification layer). 
\vspace{-5pt}
\paragraph{Block Expansion.}
We introduce the concept of Block Expansion for fine-tuning pre-trained ViTs, building upon an idea that was recently proposed for language models~\cite{wu2024llama} but has yet to be explored in vision. This technique is used to augment the capacity of a model without altering its initial output. In a ViT model comprised of sequential transformer blocks $(\phi_0, \phi_1,...,\phi_N)$, Block Expansion adds an identity block $(\phi_{id})$ after a set of transformer blocks such that $\phi_{id} (x) = x$, meaning it returns the input as its output, ensuring the model's output remains unchanged immediately after expansion. To expand a model from $N$ to $N'$ blocks, the original blocks are first grouped into sets containing $M$ blocks each. Within each set, an identity copy of the topmost block is created and placed on top, effectively increasing the model’s depth without initially changing its behavior. In each newly expanded block, two linear layers are zero-initialized to enable identity mapping, as shown in Figure~\ref{fig:framework} (c). These newly added blocks are only fine-tuned with the new data while the remaining blocks are frozen.  
\vspace{-5pt}
\paragraph{Low Rank Adaptation (LoRA).} LoRA has gained significant popularity for language tasks due to its effectiveness in fine-tuning large language models (LLMs) like GPT~\cite{brown2020language}. However, the adoption of LoRA in the vision domain has been more limited. We investigate how this technique, previously celebrated in language models, can be tailored and applied to fine-tune pretrained ViTs. We introduce LoRA to ViTs by introducing auxiliary low-rank matrices $A$ and $B$ to adapt the weight matrices of a pretrained ViT model, specifically targeting the queries $(Q)$ and values $(V)$ in the multi-headed self-attention mechanism.
In practice, for a given weight matrix $W$, the adaptation is performed by calculating $W' = W + AB$, where $W'$ is the adapted weight matrix, and $A$ and $B$ are the low-rank matrices introduced for adaptation as shown in Figure~\ref{fig:framework} (d). This modification allows the original model to retain its learned representations while gaining the flexibility to adapt to new tasks with a relatively small increase in parameters.

\begin{figure*}
    \centering
    \includegraphics[width=\linewidth]{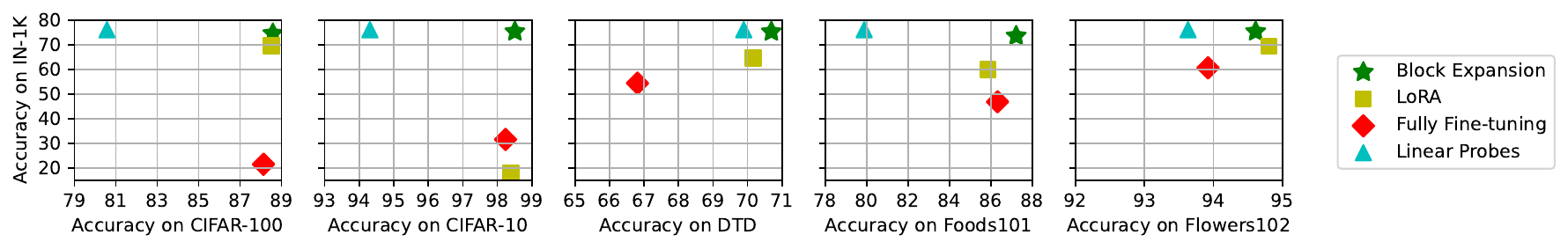}
    \caption{Comparison of top-1 accuracy between fine-tuned DINO ViT/B-16 models on transfer datasets and ImageNet-1K: the figure illustrates that models fine-tuned with Block Expansion achieve high accuracy on target datasets (e.g., CIFAR-10) while also preserving knowledge of the pre-trained dataset (ImageNet-1K).}
    \label{fig:main-res}
\end{figure*}
\section{Experiments}

\label{sec:experiments}
This section explores strategies for models to maintain performance on original tasks while excelling in new domains.
\vspace{-23pt}
\paragraph{Catastrophic Forgetting.} 
Catastrophic forgetting is when a model loses the knowledge acquired during pre-training upon being fine-tuned for a different task. To quantify this effect, we utilize the k-nearest neighbors (K-NN) method to evaluate the fine-tuned model's performance on the pre-training data. Specifically, after adding a linear classification layer to the pre-trained backbone and fine-tuning it on a transfer dataset, we apply K-NN to measure backbone's accuracy on the pre-training dataset. We compare this accuracy to the original backbone, also evaluated with K-NN on the pre-training dataset. We chose the K-NN approach for evaluating performance on ImageNet-1K (IN-1K)~\cite{deng2009imagenet} to reduce computation time.
\subsection{Main results}
In this section, we evaluate the adaptability and catastrophic forgetting of a pre-trained DINO ViT/B-16 model~\cite{caron2021emerging}, which was originally trained on the ImageNet-1K dataset. Our fine-tuning experiments are conducted on five diverse datasets: DTD (Describable Textures Dataset) \cite{Cimpoi2013DescribingTI}, Flowers102 \cite{Nilsback2008AutomatedFC}, Food101 \cite{Bossard2014Food101M}, CIFAR-10 \cite{Krizhevsky2009LearningML}, and CIFAR-100 \cite{Krizhevsky2009LearningML}. To customize the model for each dataset, we introduce an additional linear layer to the pre-trained architecture.

During the fine-tuning process, we test learning rates of 0.05, 0.01, and 0.005, and run the models for 10,000 steps, assessing accuracy every 500 steps. The checkpoint with the highest accuracy is chosen as the best model for each strategy. This model's backbone is then used for K-NN evaluation on ImageNet-1K, enabling an analysis of adaptability and the extent of catastrophic forgetting by comparing performance on ImageNet-1K and the transfer datasets.

Our findings, displayed in Figure \ref{fig:main-res}, reveal that models fine-tuned with Block Expansion typically excel in both the transfer and source domains (ImageNet-1K), effectively avoiding catastrophic forgetting. The LoRA method, although effective for CIFAR-100, shows inconsistent results and catastrophic forgetting, particularly on CIFAR-10. In our experiments, we incorporate three blocks ($p=3$)
for Block Expansion, and a rank of 8 ($r=8$) for LoRA.

We also compare two other standard fine-tuning strategies: full fine-tuning and linear probing (only linear layer fine-tuned). Full fine-tuning yields high performance on transfer datasets but leads to significant accuracy losses on ImageNet-1K, indicating a loss of original representations, aka catastrophic forgetting. Conversely, linear probing maintains good ImageNet-1K accuracy but underperforms on transfer datasets, highlighting the challenge of achieving a balance between learning new tasks and preserving existing knowledge.
\begin{figure}
    \centering
    \includegraphics[width=\linewidth]{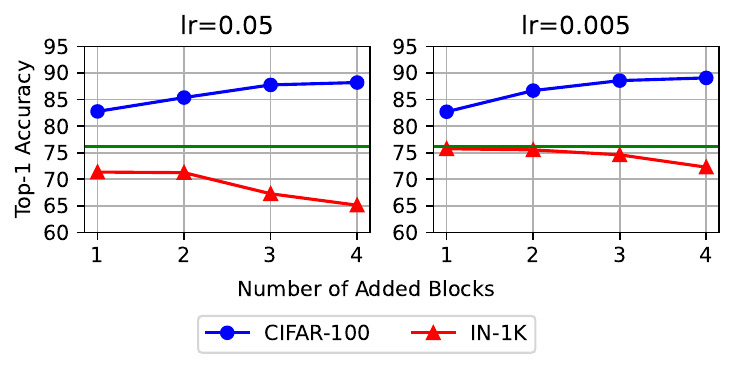}
    \caption{Exploring learning rate effects on catastrophic forgetting in fine-tuning with Block Expansion shows that higher rates, despite similar transfer dataset performance, worsen source domain forgetting, especially with more blocks added. The green line represents the accuracy of the unchanged backbone on IN-1K.}
    \label{fig:learningrate}
\end{figure}

\begin{table}[t]
\caption{Comparing fine-tuning strategies for DINO ViT/B-16 on CIFAR-100: This table presents top-1 accuracy on transfer and source datasets, alongside the count of trainable parameters. Block Expansion stands out for maximizing transfer dataset accuracy while preserving source dataset performance.}
\vspace{-12pt}
\label{ablation-table}
\begin{center}
\begin{small}
\begin{sc}
\begin{tabular*}{\linewidth}{@{\extracolsep{\fill} }l|cccc}
\toprule
\multicolumn{1}{l}{Model} & \#Params & CIFAR-100 & IN-1K & Mean \\
\toprule
\multicolumn{5}{l}{\textit{Standard Fine-tuning}} \\
\midrule
All & 85.9 M & 88.13 & 25.24 & 56.69 \\
Top-3 & 21.3 M & 84.56 & 74.15 & 79.36 \\
Linear & 76.9 K & 80.57 & \textbf{76.11} & 78.34 \\
\midrule
\multicolumn{5}{l}{\textit{LoRA}} \\
\midrule
${r=4}$ & 301 K & 87.91 & 66.82 & 77.37 \\
${r=8}$ & 448 K &  88.27 & 65.99 & 77.13 \\
${r=16}$ & 743 K &  87.84 & 65.06 & 76.45 \\
\midrule
\multicolumn{5}{l}{\textit{Block Expansion}} \\
\midrule
$p=1$ & 7.2 M & 82.72 & 75.75 & 79.24 \\
$p=2$ & 14.3 M & 86.70 & 75.54 & 81.12 \\
$p=3$ & 21.3 M & 88.58 & 74.61 & \textbf{81.60} \\
$p=4$ & 28.4 M & \textbf{89.09} & 72.28 & 80.69 \\
\bottomrule
\end{tabular*}
\end{sc}
\end{small}
\end{center}
\vspace{-20pt}
\end{table}

\subsection{Ablation Study}
In this section, we examine factors influencing the effectiveness of the above-discussed parameter-efficient fine-tuning techniques. Our focus includes the impact of the number of blocks added in Block Expansion, the rank used in LoRA, and the extent of layer fine-tuning. CIFAR-100 serves as our chosen transfer dataset for these experiments.

The results, presented in Table \ref{ablation-table}, indicate that Block Expansion leads to the highest mean accuracy compared to other fine-tuning approaches, with the optimal configuration adding three blocks ($p=3$), achieving 81.60\% mean accuracy. This performance across CIFAR-100 and IN-1K, reflects the model's ability to adapt to new domains while retaining knowledge from the source domain. However, we note a trade-off in accuracy between CIFAR-100 and IN-1K when additional blocks are added. Moreover, the data suggest that the rank parameter in LoRA does not significantly impact mean accuracy, indicating a plateau in effectiveness. Block Expansion with three blocks ($p=3$) has the same number of trainable parameters as fine-tuning the top three layers of ViT (Top 3), yet it achieves better accuracy on both CIFAR-100 and IN-1K, highlighting the effectiveness of Block Expansion. However, it is worth mentioning that the overall parameter count for Block Expansion is higher due to the addition of new layers.
\begin{figure}
    \centering
    \includegraphics[width=\linewidth]{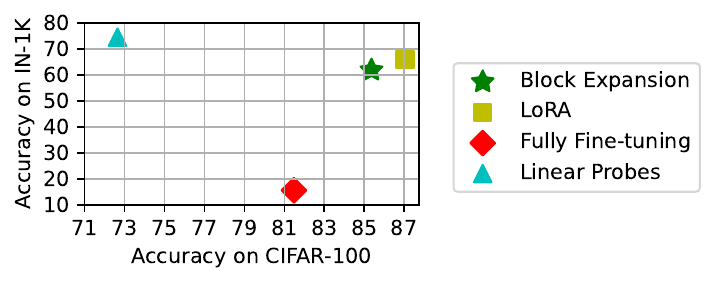}
    \vspace{-15pt}
    \caption{Comparing top-1 accuracy of fine-tuned DINO ViT/S-16 models shows LoRA and Block Expansion methods outperform traditional fine-tuning on both transfer datasets and ImageNet-1K.\vspace{-10pt}}
    \label{fig:limit}
\end{figure}

Additionally, we fine-tuned the model by varying the number of added blocks and learning rates while using the Block Expansion strategy. As depicted in Figure \ref{fig:learningrate}, while a high learning rate does not significantly affect performance on the transfer dataset, it can lead to catastrophic forgetting, with accuracy decreasing by 10\%, a phenomenon not encountered at lower learning rates. To further validate the effectiveness of the methods across different models, we applied them to a smaller DINO ViT variant. As illustrated in Figure \ref{fig:limit}, fully fine-tuning ViT-S/16 led to a reduction in ImageNet-1K accuracy to 15.77\%, whereas employing LoRA and Block Expansion maintained performance on both CIFAR-100 and ImageNet-1K.

\section{Conclusion}
\label{sec:conclusion}
In this work, we explored parameter-efficient fine-tuning (PEFT) strategies, namely Block Expansion and Low Rank Adaptaion (LoRA), to mitigate catastrophic forgetting in ViTs. We found that catastrophic forgetting, while already present in neural networks, is particularly magnified in ViTs. Our experiments show that these PEFT strategies outperform standard fine-tuning methods, achieving better performance with greater parameter efficiency. Although LoRA is generally effective for fine-tuning, it can perform poorly on simpler datasets like CIFAR-10, suggesting further investigation. Our adapted Block Expansion technique emerges as a robust method, not only enabling better fine-tuning but also retaining most of its performance on the original pre-training dataset. This indicates that Block Expansion can decrease the catastrophic forgetting faced by ViTs, allowing them to adapt to new domains while preserving their previously acquired knowledge.
\vspace{-10pt}
\paragraph{Acknowledgement.} This work utilized the Alpine high performance computing resource at the University of Colorado Boulder. Alpine is jointly funded by the University of Colorado Boulder, the University of Colorado Anschutz, Colorado State University, and the National Science Foundation (Award 2201538).

{
    \small
    \bibliographystyle{ieeenat_fullname}
    \bibliography{main}

\begin{thebibliography}{34}
\providecommand{\natexlab}[1]{#1}
\providecommand{\url}[1]{\texttt{#1}}
\expandafter\ifx\csname urlstyle\endcsname\relax
  \providecommand{\doi}[1]{doi: #1}\else
  \providecommand{\doi}{doi: \begingroup \urlstyle{rm}\Url}\fi

\bibitem[Bossard et~al.(2014)Bossard, Guillaumin, and Gool]{Bossard2014Food101M}
Lukas Bossard, Matthieu Guillaumin, and Luc~Van Gool.
\newblock Food-101 - mining discriminative components with random forests.
\newblock In \emph{European Conference on Computer Vision}, 2014.

\bibitem[Brown et~al.(2020)Brown, Mann, Ryder, Subbiah, Kaplan, Dhariwal, Neelakantan, Shyam, Sastry, Askell, et~al.]{brown2020language}
Tom Brown, Benjamin Mann, Nick Ryder, Melanie Subbiah, Jared~D Kaplan, Prafulla Dhariwal, Arvind Neelakantan, Pranav Shyam, Girish Sastry, Amanda Askell, et~al.
\newblock Language models are few-shot learners.
\newblock \emph{Advances in neural information processing systems}, 33:\penalty0 1877--1901, 2020.

\bibitem[Caron et~al.(2021)Caron, Touvron, Misra, J{\'e}gou, Mairal, Bojanowski, and Joulin]{caron2021emerging}
Mathilde Caron, Hugo Touvron, Ishan Misra, Herv{\'e} J{\'e}gou, Julien Mairal, Piotr Bojanowski, and Armand Joulin.
\newblock Emerging properties in self-supervised vision transformers.
\newblock In \emph{Proceedings of the IEEE/CVF international conference on computer vision}, pages 9650--9660, 2021.

\bibitem[Chefer et~al.(2022)Chefer, Schwartz, and Wolf]{chefer2022optimizing}
Hila Chefer, Idan Schwartz, and Lior Wolf.
\newblock Optimizing relevance maps of vision transformers improves robustness.
\newblock \emph{Advances in Neural Information Processing Systems}, 35:\penalty0 33618--33632, 2022.

\bibitem[Chen et~al.(2022)Chen, Ge, Tong, Wang, Song, Wang, and Luo]{chen2022adaptformer}
Shoufa Chen, Chongjian Ge, Zhan Tong, Jiangliu Wang, Yibing Song, Jue Wang, and Ping Luo.
\newblock Adaptformer: Adapting vision transformers for scalable visual recognition.
\newblock \emph{Advances in Neural Information Processing Systems}, 35:\penalty0 16664--16678, 2022.

\bibitem[Cimpoi et~al.(2013)Cimpoi, Maji, Kokkinos, Mohamed, and Vedaldi]{Cimpoi2013DescribingTI}
Mircea Cimpoi, Subhransu Maji, Iasonas Kokkinos, Sammy Mohamed, and Andrea Vedaldi.
\newblock Describing textures in the wild.
\newblock \emph{2014 IEEE Conference on Computer Vision and Pattern Recognition}, pages 3606--3613, 2013.

\bibitem[Deng et~al.(2009)Deng, Dong, Socher, Li, Li, and Fei-Fei]{deng2009imagenet}
Jia Deng, Wei Dong, Richard Socher, Li-Jia Li, Kai Li, and Li Fei-Fei.
\newblock Imagenet: A large-scale hierarchical image database.
\newblock In \emph{2009 IEEE conference on computer vision and pattern recognition}, pages 248--255. Ieee, 2009.

\bibitem[Dosovitskiy et~al.(2020)Dosovitskiy, Beyer, Kolesnikov, Weissenborn, Zhai, Unterthiner, Dehghani, Minderer, Heigold, Gelly, et~al.]{dosovitskiy2020image}
Alexey Dosovitskiy, Lucas Beyer, Alexander Kolesnikov, Dirk Weissenborn, Xiaohua Zhai, Thomas Unterthiner, Mostafa Dehghani, Matthias Minderer, Georg Heigold, Sylvain Gelly, et~al.
\newblock An image is worth 16x16 words: Transformers for image recognition at scale.
\newblock \emph{arXiv preprint arXiv:2010.11929}, 2020.

\bibitem[Goodfellow et~al.(2013)Goodfellow, Mirza, Xiao, Courville, and Bengio]{goodfellow2013empirical}
Ian~J Goodfellow, Mehdi Mirza, Da Xiao, Aaron Courville, and Yoshua Bengio.
\newblock An empirical investigation of catastrophic forgetting in gradient-based neural networks.
\newblock \emph{arXiv preprint arXiv:1312.6211}, 2013.

\bibitem[He et~al.(2022)He, Li, Zhang, Yang, and Wang]{He2022ParameterEfficientMA}
Xuehai He, Chunyuan Li, Pengchuan Zhang, Jianwei Yang, and Xin~Eric Wang.
\newblock Parameter-efficient model adaptation for vision transformers.
\newblock In \emph{AAAI Conference on Artificial Intelligence}, 2022.

\bibitem[Houlsby et~al.(2019)Houlsby, Giurgiu, Jastrzebski, Morrone, de~Laroussilhe, Gesmundo, Attariyan, and Gelly]{Houlsby2019Parameter}
Neil Houlsby, Andrei Giurgiu, Stanislaw Jastrzebski, Bruna Morrone, Quentin de Laroussilhe, Andrea Gesmundo, Mona Attariyan, and Sylvain Gelly.
\newblock Parameter-efficient transfer learning for nlp.
\newblock \emph{ArXiv}, abs/1902.00751, 2019.

\bibitem[Hu et~al.(2021)Hu, Shen, Wallis, Allen-Zhu, Li, Wang, Wang, and Chen]{hu2021lora}
Edward~J Hu, Yelong Shen, Phillip Wallis, Zeyuan Allen-Zhu, Yuanzhi Li, Shean Wang, Lu Wang, and Weizhu Chen.
\newblock Lora: Low-rank adaptation of large language models.
\newblock \emph{arXiv preprint arXiv:2106.09685}, 2021.

\bibitem[Jie et~al.(2023)Jie, Wang, and Deng]{Jie2023RevisitingTP}
Shibo Jie, Haoqing Wang, and Zhiwei Deng.
\newblock Revisiting the parameter efficiency of adapters from the perspective of precision redundancy.
\newblock \emph{2023 IEEE/CVF International Conference on Computer Vision (ICCV)}, pages 17171--17180, 2023.

\bibitem[Krizhevsky(2009)]{Krizhevsky2009LearningML}
Alex Krizhevsky.
\newblock Learning multiple layers of features from tiny images.
\newblock 2009.

\bibitem[Lagunas et~al.(2023)Lagunas, Impata, Martinez, Fernandez, Georgakis, Braun, and Bertrand]{lagunas2023transfer}
Manuel Lagunas, Brayan Impata, Victor Martinez, Virginia Fernandez, Christos Georgakis, Sofia Braun, and Felipe Bertrand.
\newblock Transfer learning for fine-grained classification using semi-supervised learning and visual transformers.
\newblock \emph{arXiv preprint arXiv:2305.10018}, 2023.

\bibitem[Lee et~al.(2022)Lee, Chen, Tajwar, Kumar, Yao, Liang, and Finn]{Lee2022SurgicalFI}
Yoonho Lee, Annie~S. Chen, Fahim Tajwar, Ananya Kumar, Huaxiu Yao, Percy Liang, and Chelsea Finn.
\newblock Surgical fine-tuning improves adaptation to distribution shifts.
\newblock \emph{ArXiv}, abs/2210.11466, 2022.

\bibitem[Liang et~al.(2022)Liang, Yuan, Ding, Luo, Lin, Jia, Zhang, Zhang, and Hu]{liang2022expediting}
Weicong Liang, Yuhui Yuan, Henghui Ding, Xiao Luo, Weihong Lin, Ding Jia, Zheng Zhang, Chao Zhang, and Han Hu.
\newblock Expediting large-scale vision transformer for dense prediction without fine-tuning.
\newblock \emph{Advances in Neural Information Processing Systems}, 35:\penalty0 35462--35477, 2022.

\bibitem[Loaliyan and Steeg(2024)]{Loaliyan_2024_CVPR}
Soroosh~Safari Loaliyan and Greg~Ver Steeg.
\newblock Comparative analysis of generalization and harmonization methods for 3d brain fmri images: A case study on openbhb dataset.
\newblock In \emph{Proceedings of the IEEE/CVF Conference on Computer Vision and Pattern Recognition (CVPR) Workshops}, pages 4915--4923, 2024.

\bibitem[McCloskey and Cohen(1989)]{mccloskey1989catastrophic}
Michael McCloskey and Neal~J Cohen.
\newblock Catastrophic interference in connectionist networks: The sequential learning problem.
\newblock In \emph{Psychology of learning and motivation}, pages 109--165. Elsevier, 1989.

\bibitem[Mermillod et~al.(2013)Mermillod, Bugaiska, and Bonin]{mermillod2013stability}
Martial Mermillod, Aur{\'e}lia Bugaiska, and Patrick Bonin.
\newblock The stability-plasticity dilemma: Investigating the continuum from catastrophic forgetting to age-limited learning effects.
\newblock \emph{Frontiers in psychology}, 4:\penalty0 54654, 2013.

\bibitem[Nilsback and Zisserman(2008)]{Nilsback2008AutomatedFC}
Maria-Elena Nilsback and Andrew Zisserman.
\newblock Automated flower classification over a large number of classes.
\newblock \emph{2008 Sixth Indian Conference on Computer Vision, Graphics \& Image Processing}, pages 722--729, 2008.

\bibitem[Pan et~al.(2022)Pan, Lin, Zhu, Shao, and Li]{pan2022st}
Junting Pan, Ziyi Lin, Xiatian Zhu, Jing Shao, and Hongsheng Li.
\newblock St-adapter: Parameter-efficient image-to-video transfer learning.
\newblock \emph{Advances in Neural Information Processing Systems}, 35:\penalty0 26462--26477, 2022.

\bibitem[Ramasesh et~al.(2022)Ramasesh, Lewkowycz, and Dyer]{Ramasesh2022EffectOS}
Vinay~Venkatesh Ramasesh, Aitor Lewkowycz, and Ethan Dyer.
\newblock Effect of scale on catastrophic forgetting in neural networks.
\newblock In \emph{International Conference on Learning Representations}, 2022.

\bibitem[Safaei et~al.(2023)Safaei, Vibashan, de~Melo, Hu, and Patel]{safaei2023open}
Bardia Safaei, VS Vibashan, Celso~M de Melo, Shuowen Hu, and Vishal~M Patel.
\newblock Open-set automatic target recognition.
\newblock In \emph{ICASSP 2023-2023 IEEE International Conference on Acoustics, Speech and Signal Processing (ICASSP)}, pages 1--5. IEEE, 2023.

\bibitem[Safaei et~al.(2024)Safaei, Vibashan, de~Melo, and Patel]{safaei2024entropic}
Bardia Safaei, VS Vibashan, Celso~M de Melo, and Vishal~M Patel.
\newblock Entropic open-set active learning.
\newblock In \emph{Proceedings of the AAAI Conference on Artificial Intelligence}, pages 4686--4694, 2024.

\bibitem[Sung et~al.(2021)Sung, Cho, and Bansal]{Sung2021VLADAPTERPT}
Yi-Lin Sung, Jaemin Cho, and Mohit Bansal.
\newblock Vl-adapter: Parameter-efficient transfer learning for vision-and-language tasks.
\newblock \emph{2022 IEEE/CVF Conference on Computer Vision and Pattern Recognition (CVPR)}, pages 5217--5227, 2021.

\bibitem[Volpi et~al.(2021)Volpi, Larlus, and Rogez]{volpi2021continual}
Riccardo Volpi, Diane Larlus, and Gr{\'e}gory Rogez.
\newblock Continual adaptation of visual representations via domain randomization and meta-learning.
\newblock In \emph{Proceedings of the IEEE/CVF Conference on Computer Vision and Pattern Recognition}, pages 4443--4453, 2021.

\bibitem[Wang et~al.(2023)Wang, Yang, Chang, Jin, Sun, Zhang, Luo, and Tian]{wang2023parameter}
Haixin Wang, Xinlong Yang, Jianlong Chang, Dian Jin, Jinan Sun, Shikun Zhang, Xiao Luo, and Qi Tian.
\newblock Parameter-efficient tuning of large-scale multimodal foundation model.
\newblock In \emph{Thirty-seventh Conference on Neural Information Processing Systems}, 2023.

\bibitem[Wang et~al.(2022)Wang, Mukherjee, Liu, Gao, and Gao]{Wang2022AdaMixMF}
Yaqing Wang, Subhabrata Mukherjee, Xiaodong Liu, Jing Gao, and Jianfeng Gao.
\newblock Adamix: Mixture-of-adaptations for parameter-efficient model tuning.
\newblock \emph{ArXiv}, abs/2210.17451, 2022.

\bibitem[Wu et~al.(2024)Wu, Gan, Ge, Lu, Wang, Feng, Luo, and Shan]{wu2024llama}
Chengyue Wu, Yukang Gan, Yixiao Ge, Zeyu Lu, Jiahao Wang, Ye Feng, Ping Luo, and Ying Shan.
\newblock Llama pro: Progressive llama with block expansion.
\newblock \emph{arXiv preprint arXiv:2401.02415}, 2024.

\bibitem[Xin et~al.(2024)Xin, Luo, Zhou, Du, Liu, Fan, Li, and Du]{Xin2024ParameterEfficientFF}
Yi Xin, Siqi Luo, Haodi Zhou, Junlong Du, Xiaohong Liu, Yue Fan, Qing Li, and Yuntao Du.
\newblock Parameter-efficient fine-tuning for pre-trained vision models: A survey.
\newblock \emph{ArXiv}, abs/2402.02242, 2024.

\bibitem[Yuan et~al.(2024)Yuan, Zhang, and Shan]{Yuan2024FullLoRAATEB}
Zheng Yuan, Jie Zhang, and Shiguang Shan.
\newblock Fulllora-at: Efficiently boosting the robustness of pretrained vision transformers.
\newblock \emph{ArXiv}, abs/2401.01752, 2024.

\bibitem[Zhai et~al.(2023)Zhai, Tong, Li, Cai, Qu, Lee, and Ma]{Zhai2023InvestigatingTC}
Yuexiang Zhai, Shengbang Tong, Xiao Li, Mu Cai, Qing Qu, Yong~Jae Lee, and Y. Ma.
\newblock Investigating the catastrophic forgetting in multimodal large language models.
\newblock \emph{ArXiv}, abs/2309.10313, 2023.

\bibitem[Zhang et~al.(2019)Zhang, Sax, Zamir, Guibas, and Malik]{Zhang2019SideTuningAB}
Jeffrey~O. Zhang, Alexander Sax, Amir Zamir, Leonidas~J. Guibas, and Jitendra Malik.
\newblock Side-tuning: A baseline for network adaptation via additive side networks.
\newblock In \emph{European Conference on Computer Vision}, 2019.

\end{thebibliography}
}


\end{document}